\documentclass{article}

\PassOptionsToPackage{numbers, compress}{natbib}

\usepackage[final]{neurips_2023}

\usepackage{amsmath}
\usepackage[utf8]{inputenc} %
\usepackage[T1]{fontenc}    %
\usepackage{hyperref}       %
\usepackage{cleveref}
\usepackage{url}            %
\usepackage{booktabs}       %
\usepackage{amsfonts}       %
\usepackage{nicefrac}       %
\usepackage{microtype}      %
\usepackage{xcolor}         %
\usepackage{soul}
\usepackage{graphicx}
\usepackage{booktabs}
\usepackage{multirow}

\title{CellMixer: Annotation-free Semantic Cell Segmentation of Heterogeneous Cell Populations}

\author{%
Mehdi Naouar$^{1,2,}$\thanks{Corresponding author: naouarm@cs.uni-freiburg.de} , Gabriel Kalweit$^{1,2}$, Anusha Klett$^{2,3}$, Yannick Vogt$^{1,2}$,\\\textbf{Paula Silvestrini$^2$, }\textbf{Diana Laura Infante Ramirez}$^{2,3,5}$\textbf{, Roland Mertelsmann$^{2,3}$,}\\ \textbf{Joschka Boedecker}$^{1,2,4}$\textbf{, Maria Kalweit}$^{1,2}$\\
$^1$University of Freiburg, $^2$Collaborative Research Institute Intelligent Oncology (CRIION)\\ $^3$University Medical Center Freiburg, $^4$BrainLinks-BrainTools, $^5$University of Buenos Aires\\
}

\begin{document}

\maketitle

\begin{abstract}
In recent years, several unsupervised cell segmentation methods have been presented, trying to omit the requirement of laborious pixel-level annotations for the training of a cell segmentation model. Most if not all of these methods handle the instance segmentation task by focusing on the detection of different cell instances ignoring their type. While such models prove adequate for certain tasks, like cell counting, other applications require the identification of each cell's type. In this paper, we present CellMixer, an innovative annotation-free approach for the semantic segmentation of heterogeneous cell populations. Our augmentation-based method enables the training of a segmentation model from image-level labels of homogeneous cell populations. Our results show that CellMixer can achieve competitive segmentation performance across multiple cell types and imaging modalities, demonstrating the method's scalability and potential for broader applications in medical imaging, cellular biology, and diagnostics. 
\end{abstract}

\section{Introduction}

Accurate and scalable semantic segmentation of cells in medical imaging is a key challenge with far-reaching implications for diagnostics, research, and personalized medicine. In recent years, deep learning-based methods have gained increasing traction in this fast-moving field. By improving the real-time identification of malignant cells, clinicians can benefit from the objective insights of a computational system, providing a clearer lens to observe and evaluate potential discrepancies or nuances that may be missed in manual assessments. In addition, these technologies could help improve treatment strategies by differentiating resistant cells from non-resistant ones and discerning intricate stages of cancer, such as the blast stage in leukemia. Computational tools appear particularly relevant in cases such as the identification of leukemic cells within heterogeneous populations. In the field of cell segmentation, a diverse landscape of research has emerged, ranging from the segmentation of cell instances \cite{ASHA2023105704,Falk2019} to the semantic segmentation of cell classes \cite{Sadanandan2017, Saleem2022}. While traditional supervised methods have clear advantages, they face inherent challenges, one of which is the need for manual annotation. Manual processes remain time-intensive and prone to errors, even with advanced annotation tools. For instance, the annotation of peripheral blood mononuclear cells (PBMCs) isolated from healthy donors poses its own unique difficulties. Their inherent morphological diversity and close resemblance to other cell types make manual annotation of PBMCs particularly labor-intensive and error-prone or even require fluorescent immunophenotyping markers to label the different groups within PBMCs, underscoring the need for more annotation-independent computational tools. In contrast, unsupervised learning approaches perform well on more straightforward tasks such as foreground segmentation \cite{10.1007/978-3-030-87193-2_27,10.1007/978-3-031-09037-0_20}. Yet, when faced with more complex challenges, they often lag behind the results achieved by supervised methods. This highlights the delicate balance of reducing the level of supervision while striving to maintain accuracy. In this landscape, weakly-supervised learning emerges as a compelling middle ground, aiming to balance the benefits of both approaches. %

\begin{figure}[t]
    \centering
    \includegraphics[width=0.96\textwidth]{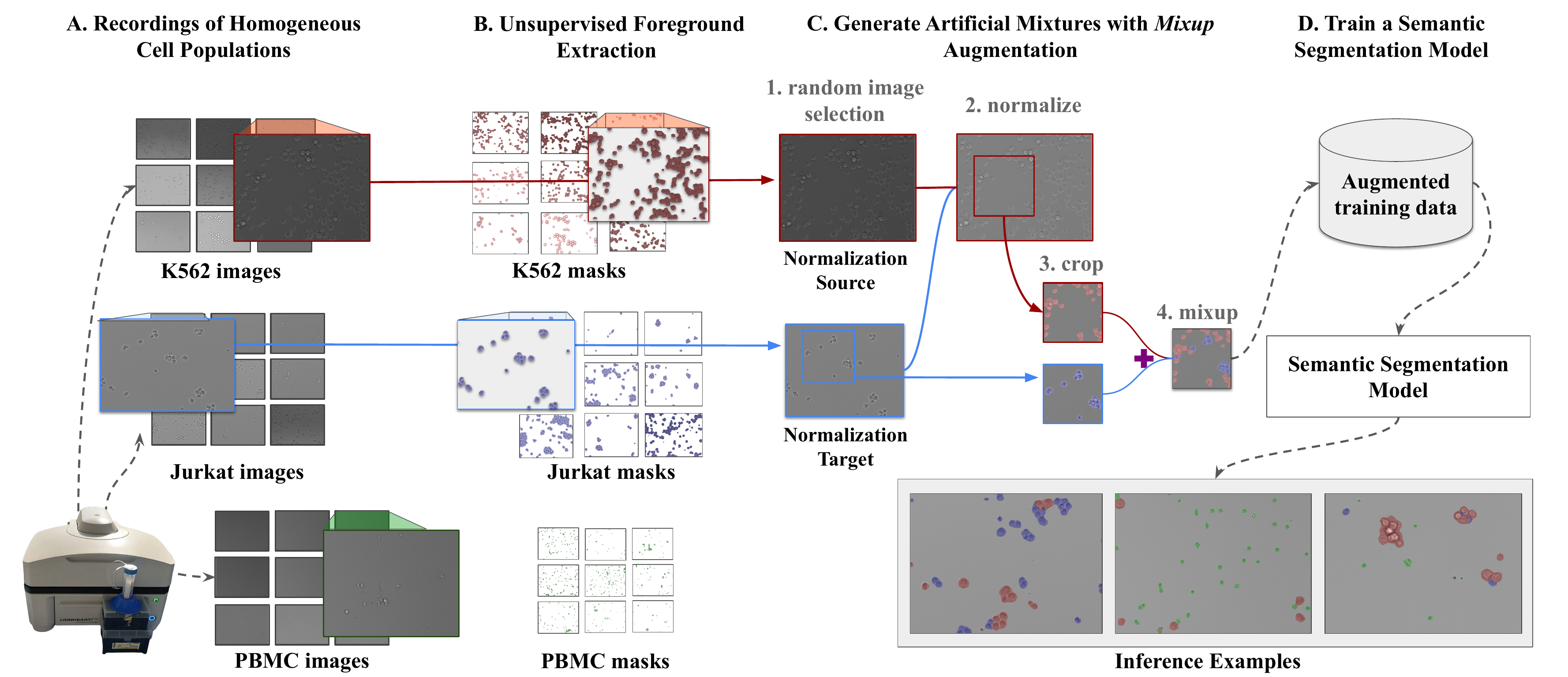}
    \caption{CellMixer Pipeline: (A) Homogeneous Cell Populations are recorded separately with the Lionheart microscope. (B) Foreground annotations are extracted using an unsupervised gradient-based approach and the image label is assigned to each cell. (C) Artificial mixtures are generated using mixup augmentation. (D) A semantic segmentation model is trained from the artificial mixtures.}
    \label{fig:cellmixeroverview}
\end{figure}

In this work, we introduce CellMixer, a novel augmentation-based strategy aiming to suppress the need for labor-intensive manual annotations. Our weakly-supervised semantic cell segmentation method uses unannotated labeled brightfield images of homogeneous cell populations captured on a Lionheart automated microscope at 20x magnification and comprises four stages (cf. \Cref{fig:cellmixeroverview}). First, we record single-class populations individually. Second, we employ an unsupervised foreground segmentation method to extract single-class clumps of cells. Third, we randomly select pairs of images and normalize their brightness and saturation. Fourth, we employ mixup augmentation \cite{mixup} to generate artificially annotated mixed populations. This generated dataset of artificial mixtures is then used to train a supervised segmentation model. In our experiments, we applied this approach to simulate realistic clinical conditions by generating artificial mixtures of two different leukemic cell lines, Jurkat and K562, with PBMCs and validated our model on real mixtures of these cells.

\section{CellMixer}

\paragraph{Unsupervised Foreground Extraction}
After recording homogeneous populations, we first extract cells from their respective background as a first step towards artificially mixed populations. At its core, our unsupervised foreground segmentation method relies on the inherent contrasts and structures present in such images to distinguish regions of interest from the surrounding environment. Each image $I$ is processed to accentuate its salient structures via the Sobel operators $S_x$ and $S_y$,
which effectively capture the edges, emphasizing transitions in intensity, which often correspond to cell areas in microscopy images. The gradient magnitude of the pixels in each image is then calculated as $G = \sqrt{S_x^2 + S_y^2}$. Smoothing, applied before and after the Sobel operation, provides attenuation of high-frequency noise. This helps to better highlight true cell boundaries. Subsequent refinement by morphological erosion, defined as: $(G \ominus B)(x,y) = \min_{(i,j) \in B} G(x-i, y-j)$,
where $B$ is a structuring element for the local neighborhood, helps break small intercellular connections and suppress unimportant structures, ensuring that larger, cohesive structures (cells or cell clusters) are preserved and emphasized. By (adaptively) thresholding the processed gradient magnitudes, a clear segmentation between the cells and the background is achieved. Next, the given image label is assigned to each of its foreground pixels. 

\paragraph{Artificial Mixtures}Within the training process, two image crops $I_1$ and $I_2$, along with their respective ground truth maps $M_1$ and $M_2$, are randomly selected from the dataset. These are subsequently normalized based on the mean and standard deviation of their background pixels. This step is crucial to ensure that the brightness and saturation levels of the images align properly before their combination. Note that the mean and standard deviation of all pixels of an image depend on its content, such as its foreground portion. Consequently, normalizing based on the background pixels often provides a more reliable reference for achieving the desired brightness and saturation in the resulting image.  Finally, the cell mixing operation combines both images with  $\tilde I = \lambda I_1 + (1 - \lambda) I_2$, where we empirically set $\lambda = 0.5$. The same operation is then applied to their respective ground truth maps using $\tilde M = M_2 + (1 -  \text{sign}(M_2)) \cdot M_1 $, conditionally overwriting $M_1$ with $M_2$.
\paragraph{Model Training}The resulting composite images serve as a robust training set for a semantic segmentation model.
This methodology significantly reduces the necessity for manual annotation, thereby expediting the training process and enhancing scalability.

\section{Results and Discussion}
In our experiments,  we compare a segmentation model trained on images from CellMixer (CellMixer) to the baseline which was trained on the original homogeneous cell cultures and their generated ground truth maps (Baseline). Both models are evaluated on four different datasets: the first consisting of a set of images of homogeneous cell cultures, the second of artificially generated heterogeneous cell cultures using CellMixer, and two sets of real heterogeneous cell cultures partially annotated by an expert. More details about the datasets and the experimental setup can be found in \Cref{exp_setting}.
\begin{table*}
	\centering
\caption{Comparison between the segmentation performance of CellMixer with the baseline on homogeneous cell cultures (UnMixed), artificially generated cell mixtures (Artificial Mix), real jurkat+K562 cell mixtures, and a selection of PBMC+jurkat and PBMC+K562 cell mixtures (Real PJ+PK). The  mean accuracy (mAcc) and mean Intersection over Union (mIoU) are reported.}
	\begin{tabular}{rrrrrrrrrrrr}
      \toprule
		 && \multicolumn{2}{c}{Unmixed}& \multicolumn{2}{c}{Artificial Mix} & \multicolumn{2}{c}{Real JK Mix}&\multicolumn{2}{c}{Real PJ+PK Mix}\\
		Model&Class
		& \multicolumn{1}{c}{mAcc}
		& \multicolumn{1}{c}{mIoU}
		& \multicolumn{1}{c}{mAcc}
		& \multicolumn{1}{c}{mIoU}
		&\multicolumn{1}{c}{mAcc}
		& \multicolumn{1}{c}{mIoU}&\multicolumn{1}{c}{mAcc}
		& \multicolumn{1}{c}{mIoU}\\
		\midrule

		Baseline & Backgr. & 99.11 & 98.74 & 99.32& 96.60 & - & -& - & -\\
                & Jurkat  &  93.98&87.83 & 72.79 & 64.94 & 62.04&52.18& 63.29 & 46.09 \\
                & K562  & 97.91& 92.71& 88.20 & 80.44& 71.92& 49.99 & 34.99 & 34.99\\
                &PBMC & 96.17 & 78.62 & 56.19 &39.43 &- &-& - & -\\
    \midrule
		CellMixer& Backgr.& 98.72 & 98.42& 98.12 & 97.19& - & -& - & -   \\
                & Jurkat &93.09  & 82.33 & \bf{90.41} &\bf{83.07} & \bf{76.57}& \bf{72.65}& \bf{95.76}& \bf{66.26}\\
                & K562&96.57 &  89.63 & \bf{96.68}& \bf{89.87}&\bf{90.86}& \bf{71.99}& \bf{63.70} & \bf{63.70}\\
                &PBMC &  95.0 & 69.15 &\bf{91.34} &\bf{72.28} &- & -& - &   -\\
    \bottomrule

	\end{tabular}
	\label{tab1}
\end{table*}
The results reported in \Cref{tab1} show that our method clearly outperforms the baseline on both artificial and real cell mixtures. In particular, the quantitative scores suggest that the baseline achieves a decent performance on unmixed cells, but struggles to segment cell mixtures. Based on the observations from our qualitative analysis (cf. \Cref{subseq:qualitative}), it seems apparent that this effect is due to the model's tendency to assign the same label to nearby foreground pixels. This bias, caused by the fact that the training set of the baseline contains only homogeneous cell mixtures, becomes problematic as soon as the network encounters heterogeneous cell cultures. While CellMixer significantly reduces this bias, the network still tends to assign the same label to cells in the same neighborhood in some rare cases. In addition, the qualitative analysis shows that the most challenging cases for CellMixer occur with very dense cell clusters. However, these are extremely difficult to analyze even for experts due to the limited visibility of occluded cells in the 2D projection.

\section{Conclusion}
We presented CellMixer, a novel augmentation-based weakly supervised semantic cell segmentation approach.
By suppressing the need for annotations, CellMixer greatly reduces the data generation effort by the experts and the number of mislabeled cells, as it is way easier to label homogeneous cell cultures than to annotate every single cell of a heterogeneous culture. To better cover overlapping cells, moving to multi-label segmentation and the generation of more realistic clumps of cells are promising avenues for future work.

\section*{Acknowledgments}
This project was funded by the Mertelsmann Foundation. This work is part of BrainLinks-BrainTools which is funded by the Federal Ministry of Economics, Science and Arts of Baden-Württemberg within the sustainability program for projects of the excellence initiative II.

\bibliographystyle{plainnat}
\bibliography{references}

\begin{thebibliography}{8}
\providecommand{\natexlab}[1]{#1}
\providecommand{\url}[1]{\texttt{#1}}
\expandafter\ifx\csname urlstyle\endcsname\relax
  \providecommand{\doi}[1]{doi: #1}\else
  \providecommand{\doi}{doi: \begingroup \urlstyle{rm}\Url}\fi

\bibitem[Asha et~al.(2023)Asha, Gopakumar, and Subrahmanyam]{ASHA2023105704}
S.B. Asha, G.~Gopakumar, and Gorthi R.K.~Sai Subrahmanyam.
\newblock Saliency and ballness driven deep learning framework for cell segmentation in bright field microscopic images.
\newblock \emph{Engineering Applications of Artificial Intelligence}, 118:\penalty0 105704, 2023.
\newblock ISSN 0952-1976.
\newblock \doi{https://doi.org/10.1016/j.engappai.2022.105704}.
\newblock URL \url{https://www.sciencedirect.com/science/article/pii/S0952197622006947}.

\bibitem[Falk et~al.(2019)Falk, Mai, Bensch, {\c{C}}i{\c{c}}ek, Abdulkadir, Marrakchi, B{\"o}hm, Deubner, J{\"a}ckel, Seiwald, Dovzhenko, Tietz, Dal~Bosco, Walsh, Saltukoglu, Tay, Prinz, Palme, Simons, Diester, Brox, and Ronneberger]{Falk2019}
Thorsten Falk, Dominic Mai, Robert Bensch, {\"O}zg{\"u}n {\c{C}}i{\c{c}}ek, Ahmed Abdulkadir, Yassine Marrakchi, Anton B{\"o}hm, Jan Deubner, Zoe J{\"a}ckel, Katharina Seiwald, Alexander Dovzhenko, Olaf Tietz, Cristina Dal~Bosco, Sean Walsh, Deniz Saltukoglu, Tuan~Leng Tay, Marco Prinz, Klaus Palme, Matias Simons, Ilka Diester, Thomas Brox, and Olaf Ronneberger.
\newblock U-net: deep learning for cell counting, detection, and morphometry.
\newblock \emph{Nature Methods}, 16\penalty0 (1):\penalty0 67--70, Jan 2019.
\newblock ISSN 1548-7105.
\newblock \doi{10.1038/s41592-018-0261-2}.
\newblock URL \url{https://doi.org/10.1038/s41592-018-0261-2}.

\bibitem[Han and Yin(2021)]{10.1007/978-3-030-87193-2_27}
Liang Han and Zhaozheng Yin.
\newblock Unsupervised network learning for cell segmentation.
\newblock In Marleen de~Bruijne, Philippe~C. Cattin, St{\'e}phane Cotin, Nicolas Padoy, Stefanie Speidel, Yefeng Zheng, and Caroline Essert, editors, \emph{Medical Image Computing and Computer Assisted Intervention -- MICCAI 2021}, pages 282--292, Cham, 2021. Springer International Publishing.
\newblock ISBN 978-3-030-87193-2.

\bibitem[Krug and Rohr(2022)]{10.1007/978-3-031-09037-0_20}
Carola Krug and Karl Rohr.
\newblock Unsupervised cell segmentation in fluorescence microscopy images via self-supervised learning.
\newblock In Moun{\^i}m El~Yacoubi, Eric Granger, Pong~Chi Yuen, Umapada Pal, and Nicole Vincent, editors, \emph{Pattern Recognition and Artificial Intelligence}, pages 236--247, Cham, 2022. Springer International Publishing.
\newblock ISBN 978-3-031-09037-0.

\bibitem[Sadanandan et~al.(2017)Sadanandan, Ranefall, Le~Guyader, and W{\"a}hlby]{Sadanandan2017}
Sajith~Kecheril Sadanandan, Petter Ranefall, Sylvie Le~Guyader, and Carolina W{\"a}hlby.
\newblock Automated training of deep convolutional neural networks for cell segmentation.
\newblock \emph{Scientific Reports}, 7\penalty0 (1):\penalty0 7860, Aug 2017.
\newblock ISSN 2045-2322.
\newblock \doi{10.1038/s41598-017-07599-6}.
\newblock URL \url{https://doi.org/10.1038/s41598-017-07599-6}.

\bibitem[Saleem et~al.(2022)Saleem, Amin, Sharif, Anjum, Iqbal, and Wang]{Saleem2022}
Saba Saleem, Javeria Amin, Muhammad Sharif, Muhammad~Almas Anjum, Muhammad Iqbal, and Shui-Hua Wang.
\newblock A deep network designed for segmentation and classification of leukemia using fusion of the transfer learning models.
\newblock \emph{Complex {\&} Intelligent Systems}, 8\penalty0 (4):\penalty0 3105--3120, Aug 2022.
\newblock ISSN 2198-6053.
\newblock \doi{10.1007/s40747-021-00473-z}.
\newblock URL \url{https://doi.org/10.1007/s40747-021-00473-z}.

\bibitem[Strudel et~al.(2021)Strudel, Pinel, Laptev, and Schmid]{segmenter}
Robin Strudel, Ricardo~Garcia Pinel, Ivan Laptev, and Cordelia Schmid.
\newblock Segmenter: Transformer for semantic segmentation.
\newblock In \emph{{ICCV}}, pages 7242--7252. {IEEE}, 2021.

\bibitem[Zhang et~al.(2018)Zhang, Ciss{\'{e}}, Dauphin, and Lopez{-}Paz]{mixup}
Hongyi Zhang, Moustapha Ciss{\'{e}}, Yann~N. Dauphin, and David Lopez{-}Paz.
\newblock mixup: Beyond empirical risk minimization.
\newblock In \emph{{ICLR} (Poster)}. OpenReview.net, 2018.

\end{thebibliography}

\appendix

\section{Appendix}
\subsection{Experimental Setting}
\label{exp_setting}
\subsubsection{Dataset}
The dataset consists of labeled brightfield images of homogeneous cell populations captured on a Lionheart automated microscope at 20x magnification resulting in grayscale images with a resolution of  $1224 \times 904$ pixels. The dataset is highly imbalanced; 2405 of the images are labeled as \emph{Jurkat cells}, 1603 as \emph{K562} cells, and 588 as \emph{PBMC cells}. The entire dataset is split by withholding 10\% of randomly selected images from each cell culture for validation and using the remaining images for training.

For Testing, a dataset consisting of 152 partially annotated images from heterogeneous cell mixtures is used for the validation of our method in a real setting. Note that an exhaustive accurate annotation of a heterogeneous cell culture is challenging even for an expert.

\subsubsection{Training Pipeline}
We train a deep segmentation model composed of a dinov2 encoder and a Segmenter decoder \cite{segmenter}. The model is trained for 20k Iterations with a batch size of 16 and a crop size of 518.

During training, we apply random flipping, random blurring, random Gaussian noise as well as a random modification of the brightness and contrast of the input image as data augmentation and train our model using the SGD optimizer and the Tversky loss to address the issue of data imbalance. We train our model using a learning rate of 0.001 and without weight decay=$0.0$.

The inference of the model is performed using a sliding window of size corresponding to the model input size (518).

\subsection{Qualitative Examples}
\label{subseq:qualitative}
\begin{figure}
	\centering
	\begin{tabular}{ cccc }

	 &Ground Truth & Baseline & Cellmixer  \\

	&\includegraphics[width=0.29\textwidth]{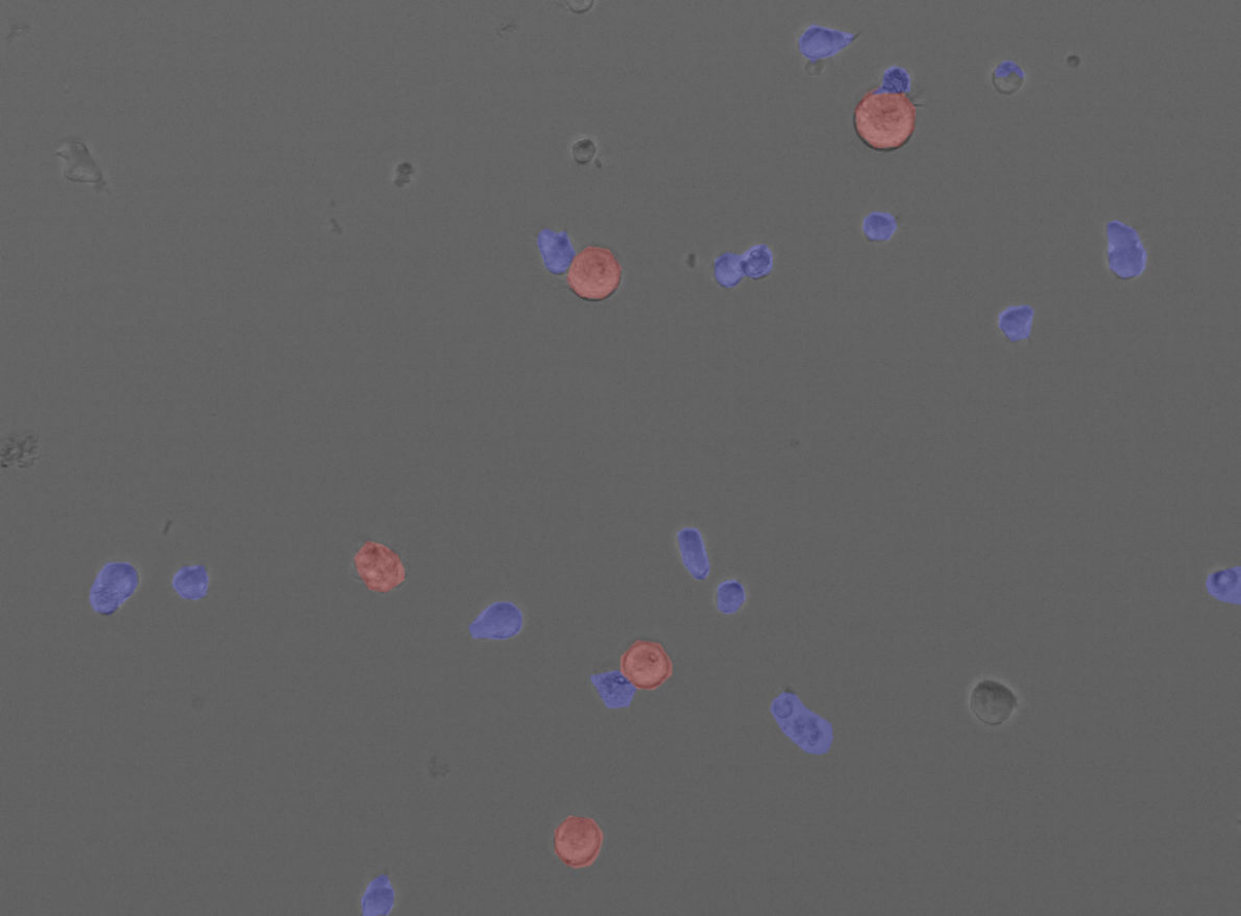}& \includegraphics[width=0.29\textwidth]{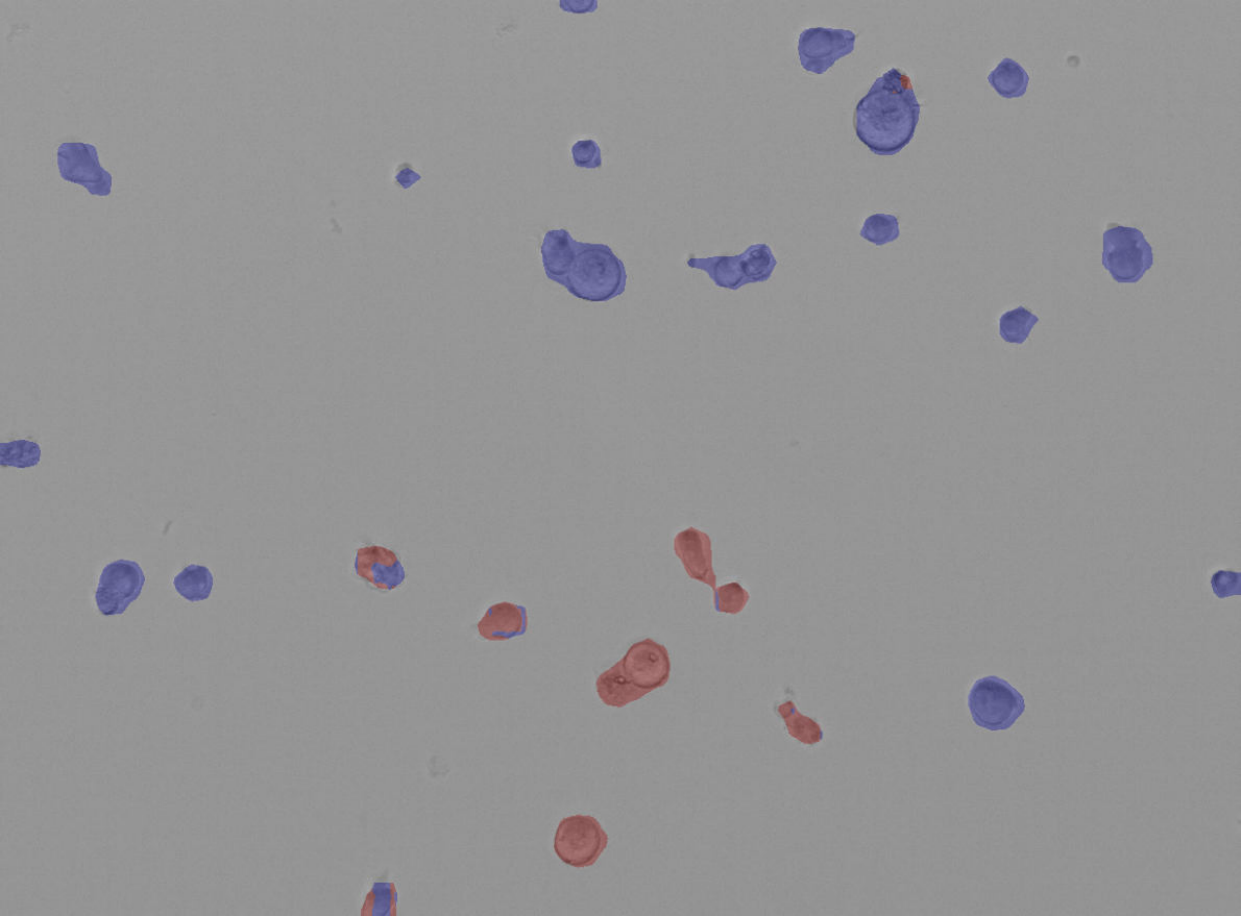}&
 \includegraphics[width=0.29\textwidth]{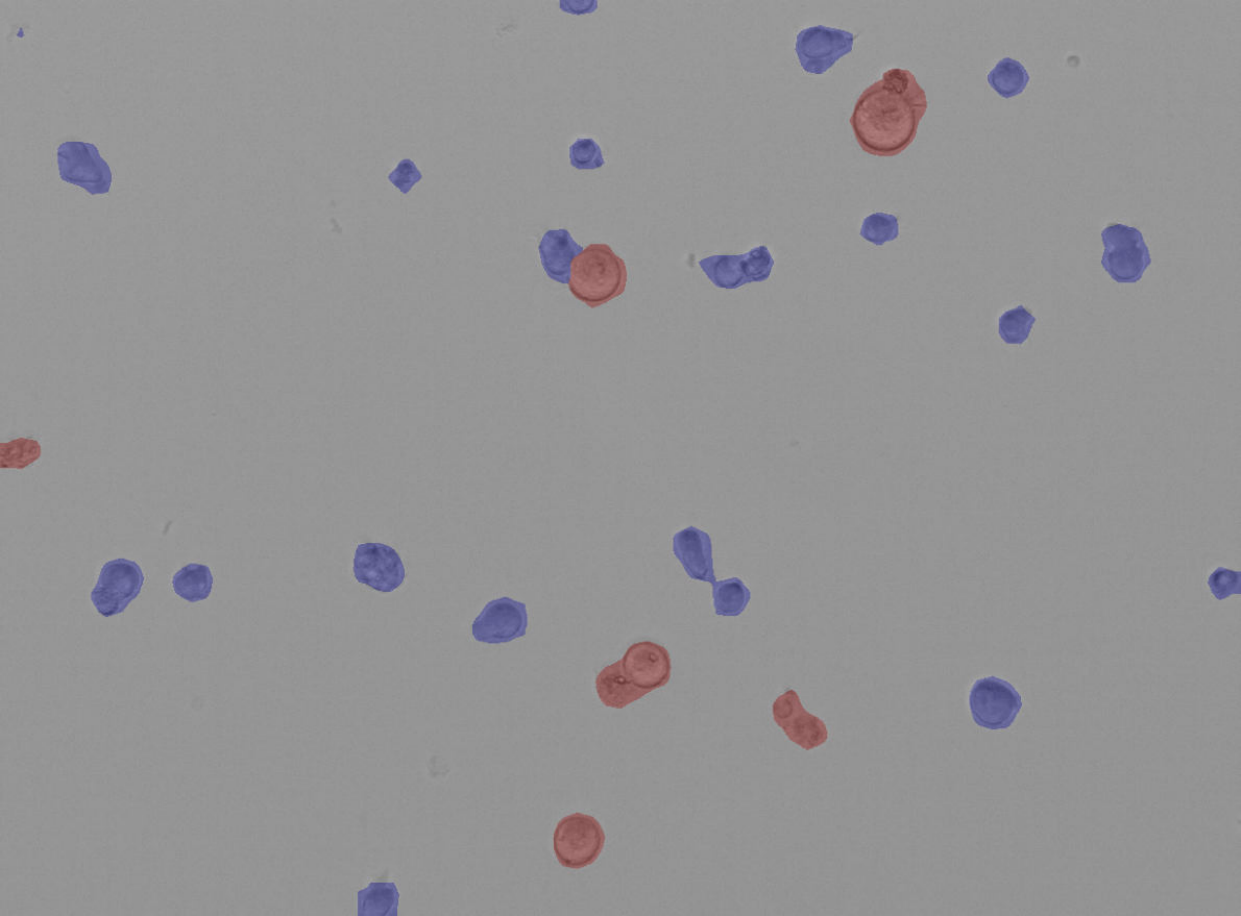}\vspace{-1.7cm} \\(1) \vspace{1.3cm}\\
 	&\includegraphics[width=0.29\textwidth]{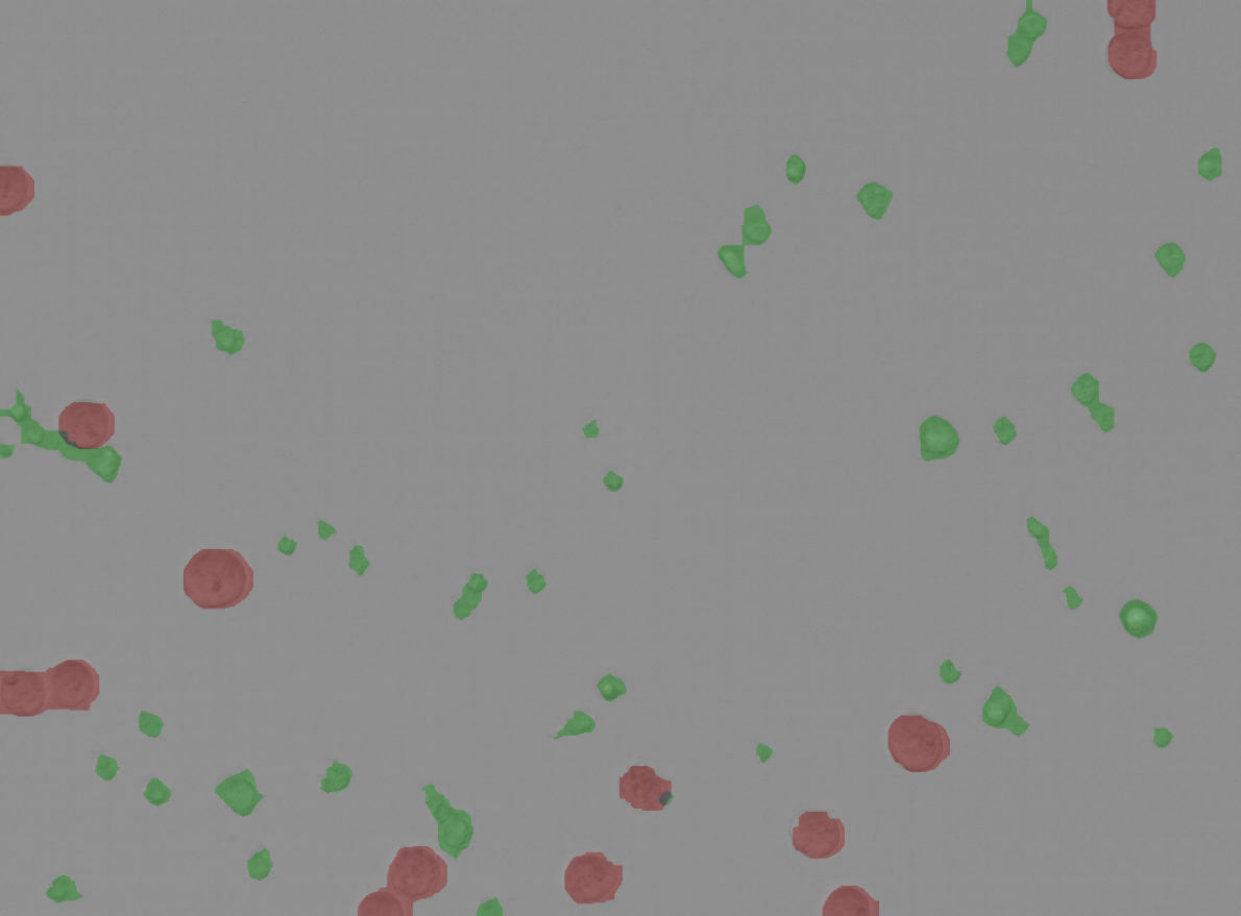}& \includegraphics[width=0.29\textwidth]{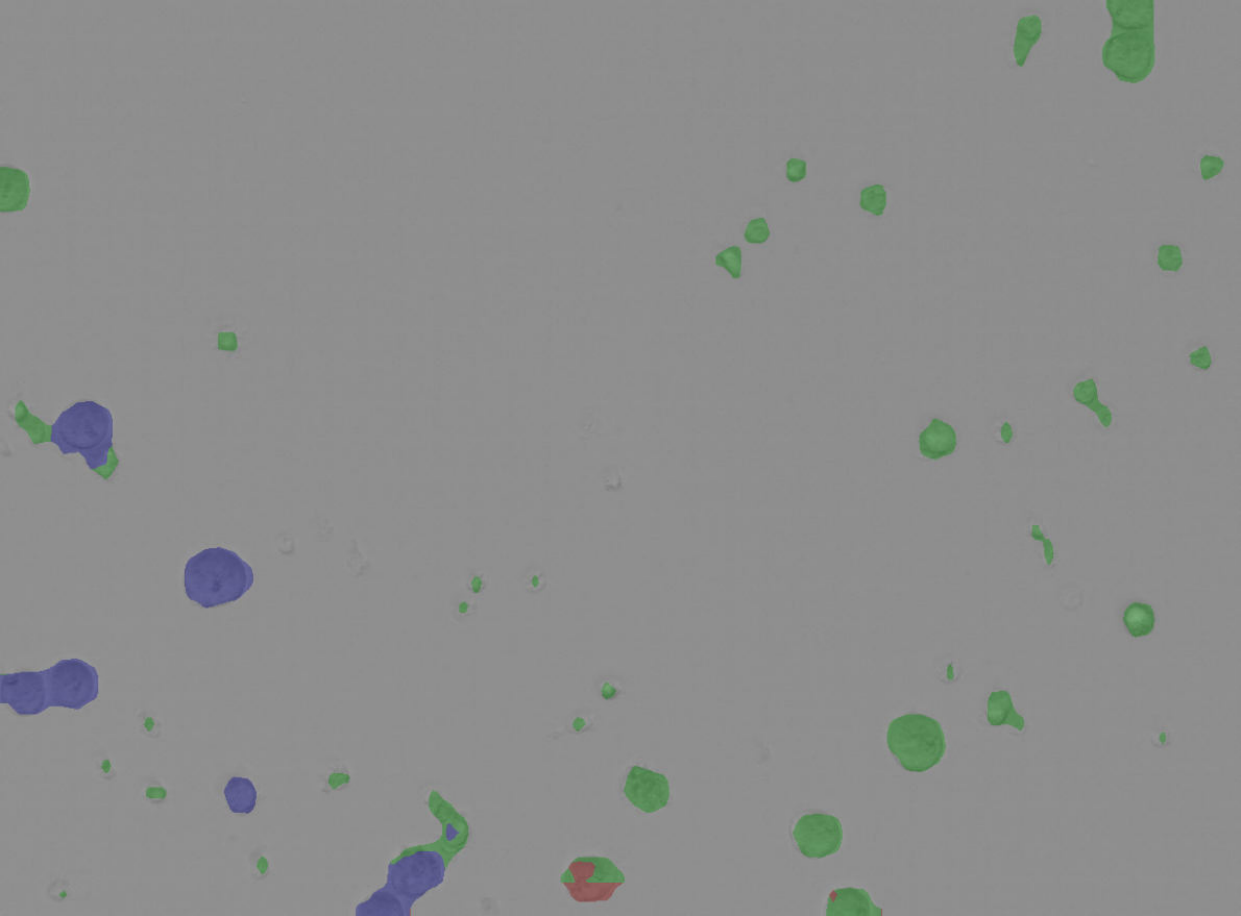}&
    \includegraphics[width=0.29\textwidth]{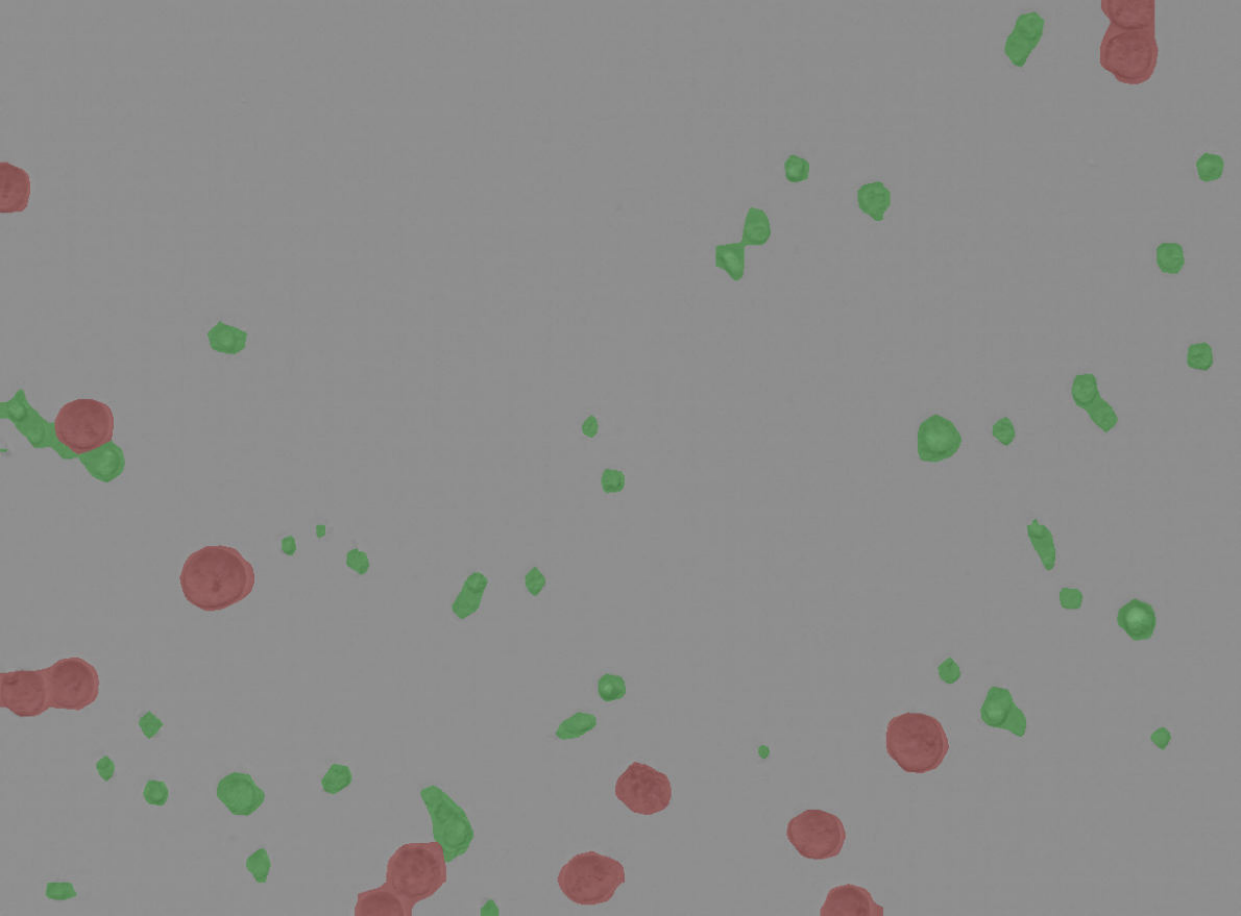} \vspace{-1.7cm} \\(2) \vspace{1.3cm}\\&\includegraphics[width=0.29\textwidth]{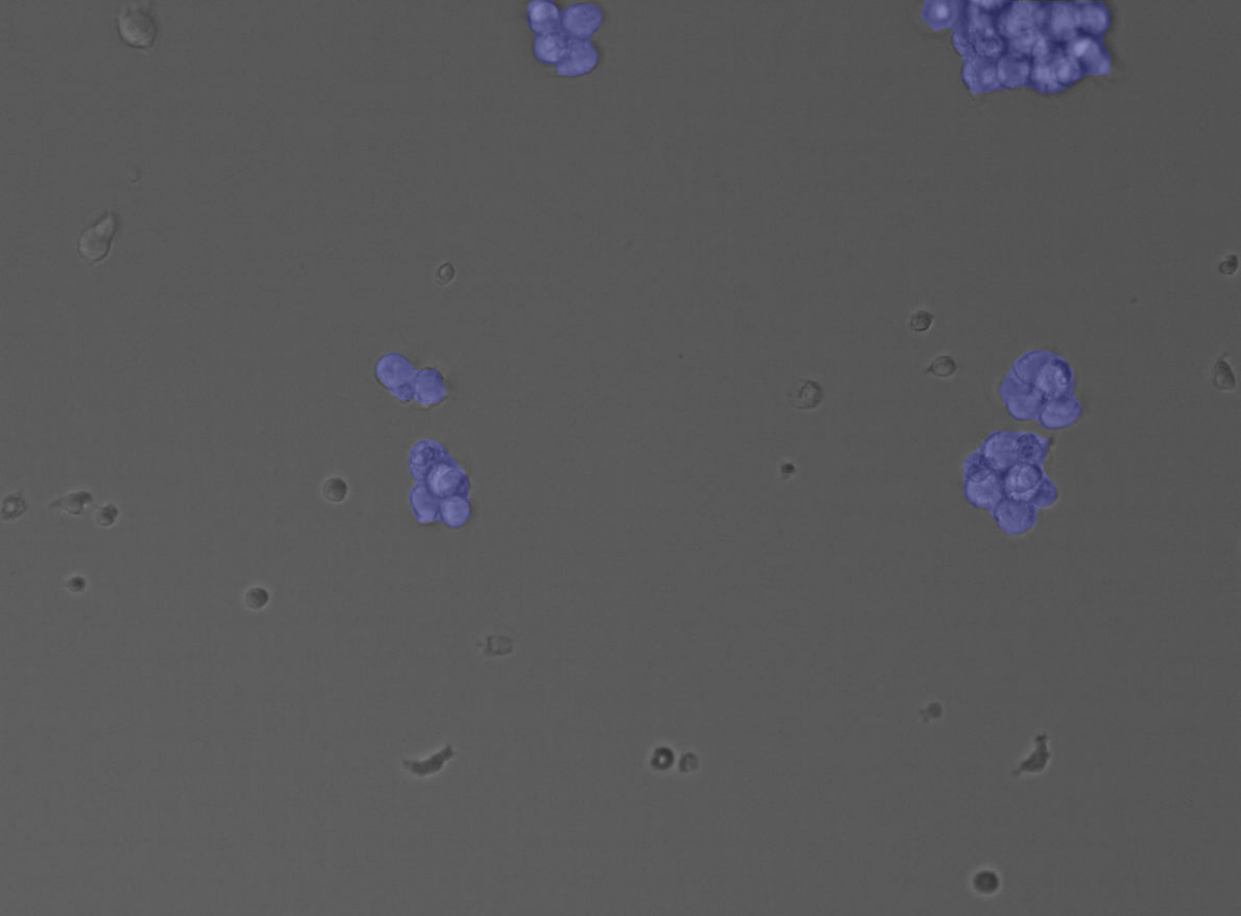}& \includegraphics[width=0.29\textwidth]{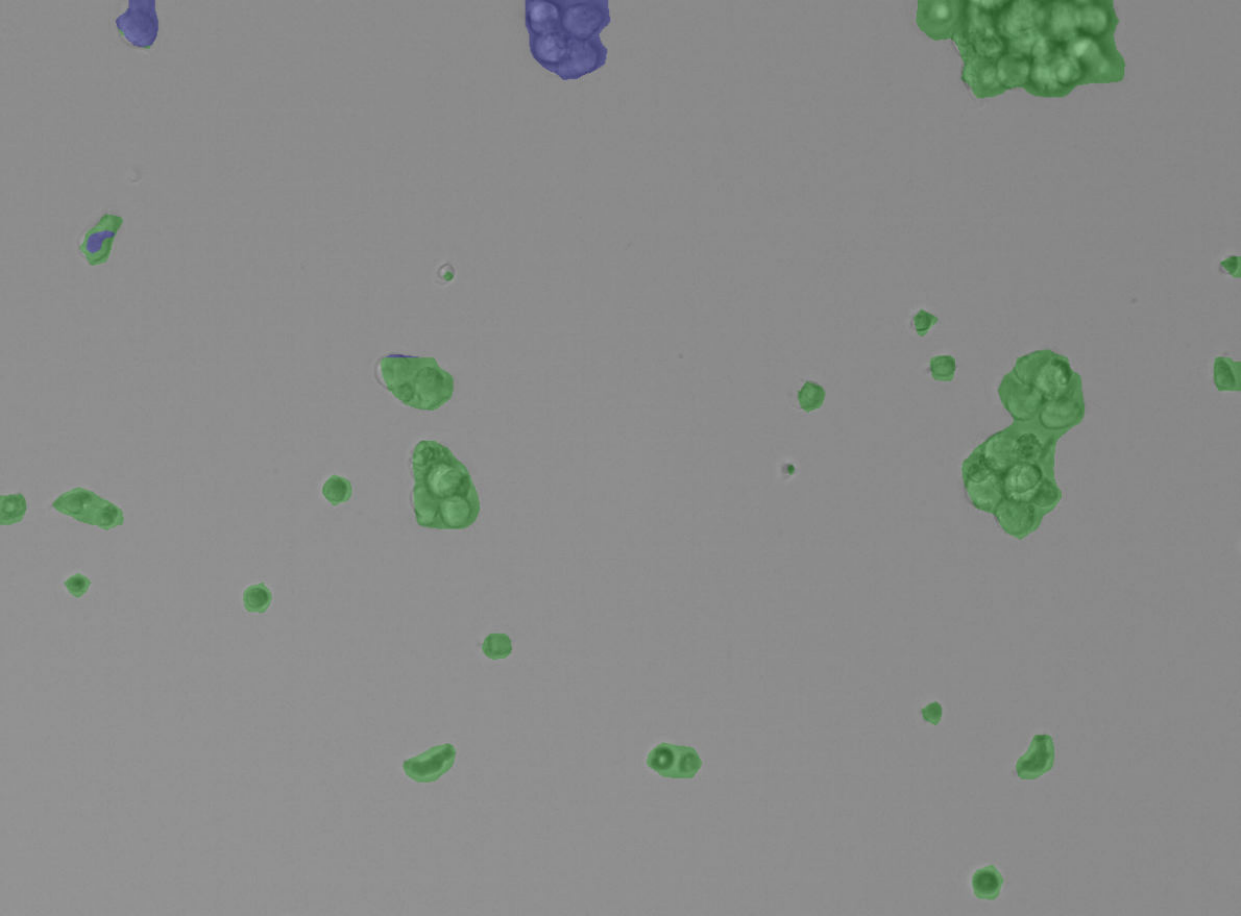}&
    \includegraphics[width=0.29\textwidth]{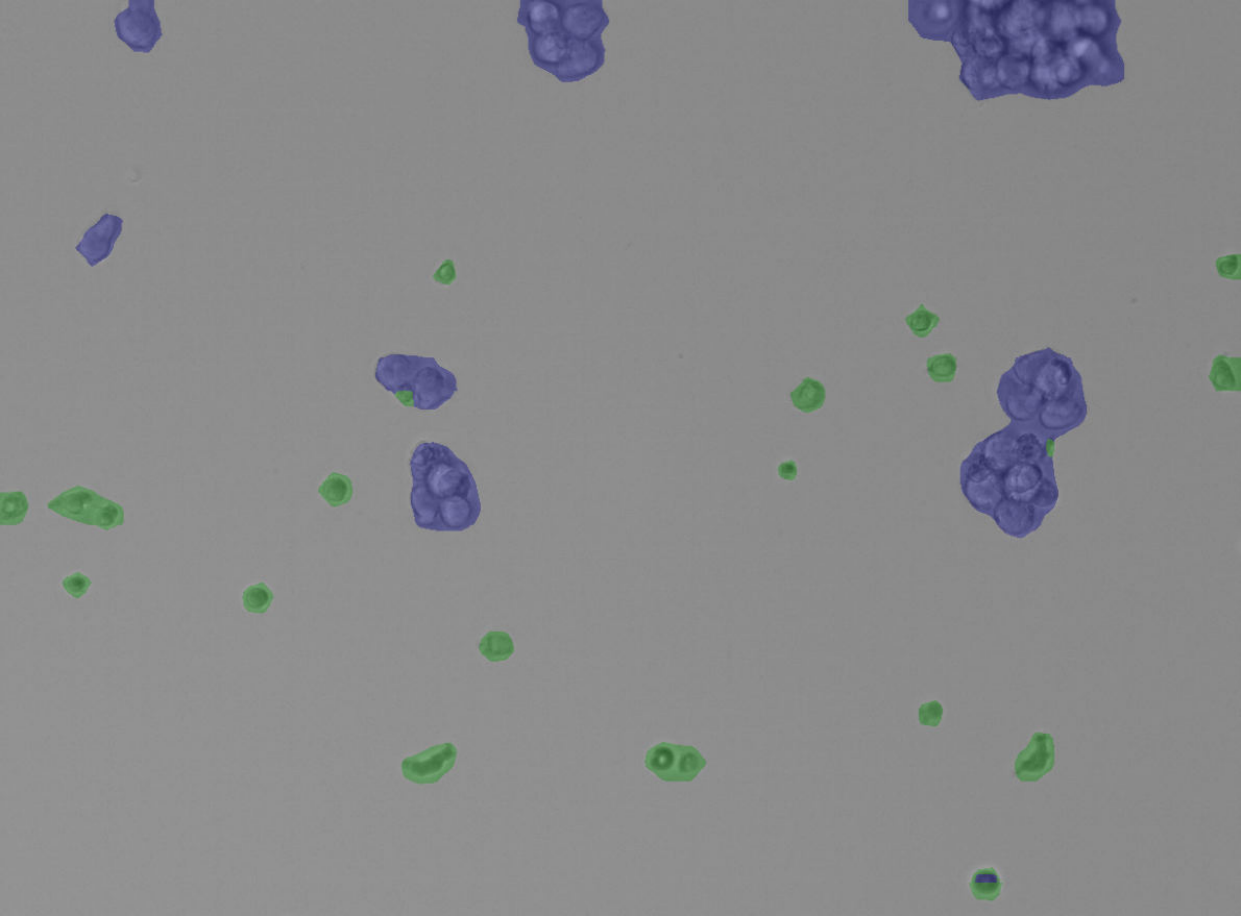} \vspace{-1.7cm} \\(3) \vspace{1.3cm}\\&\includegraphics[width=0.29\textwidth]{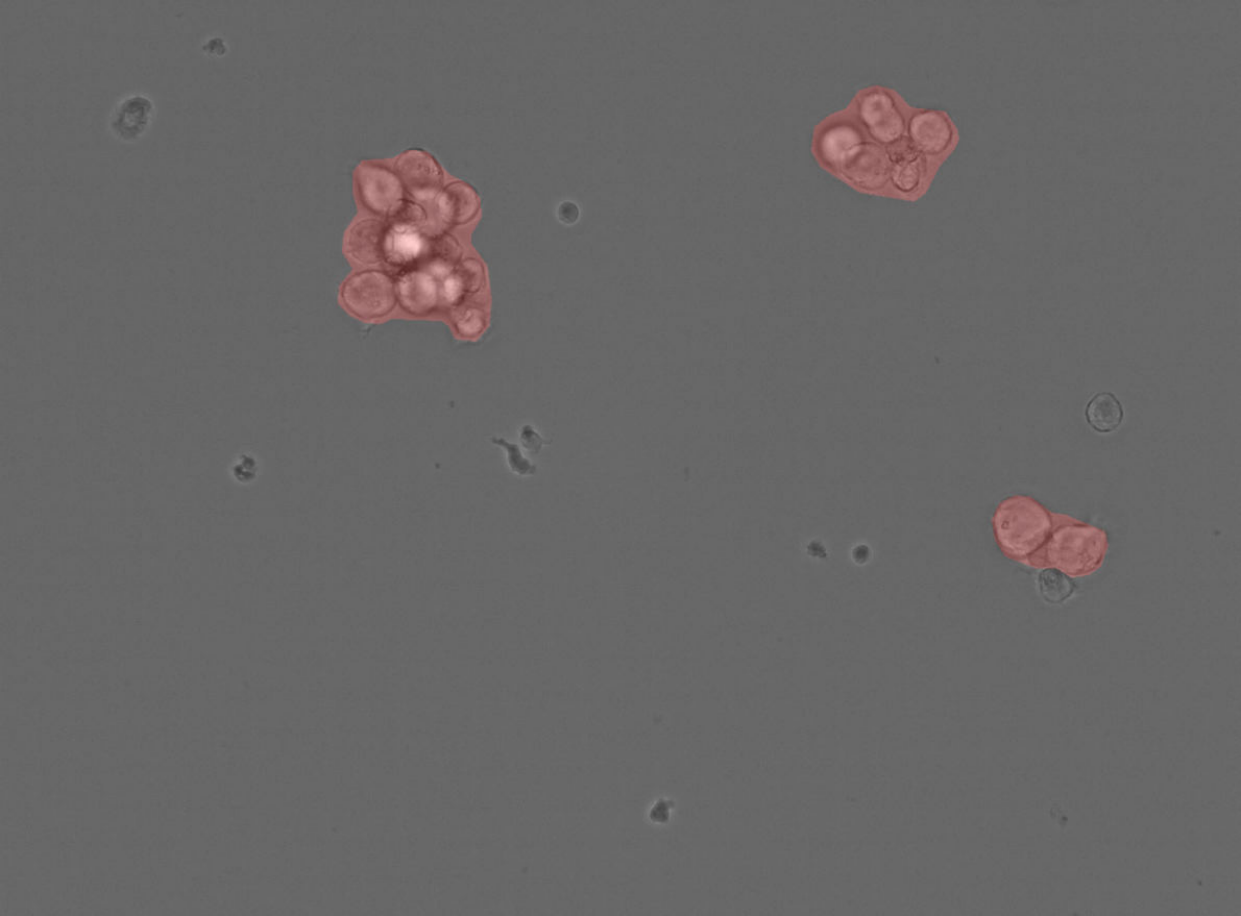}& \includegraphics[width=0.29\textwidth]{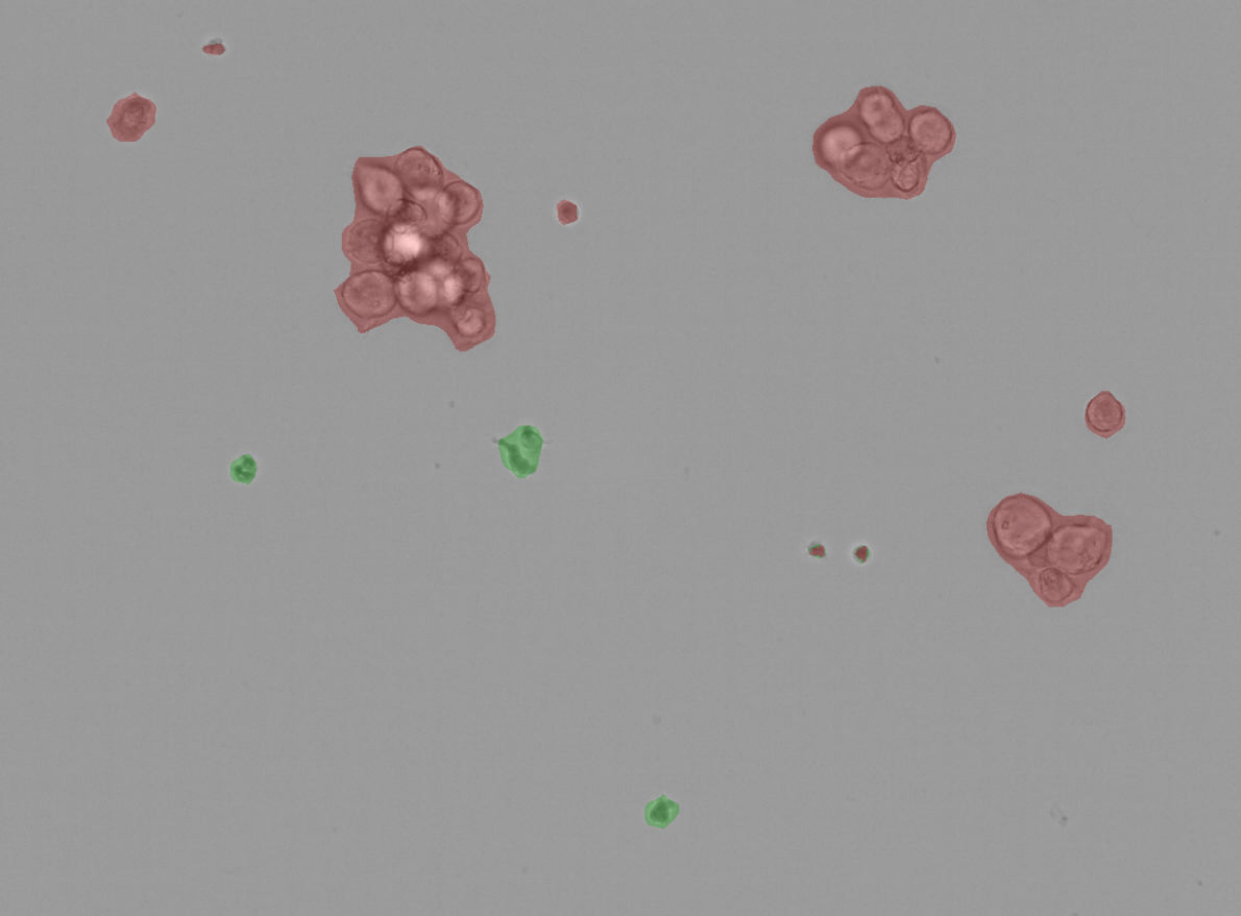}&
    \includegraphics[width=0.29\textwidth]{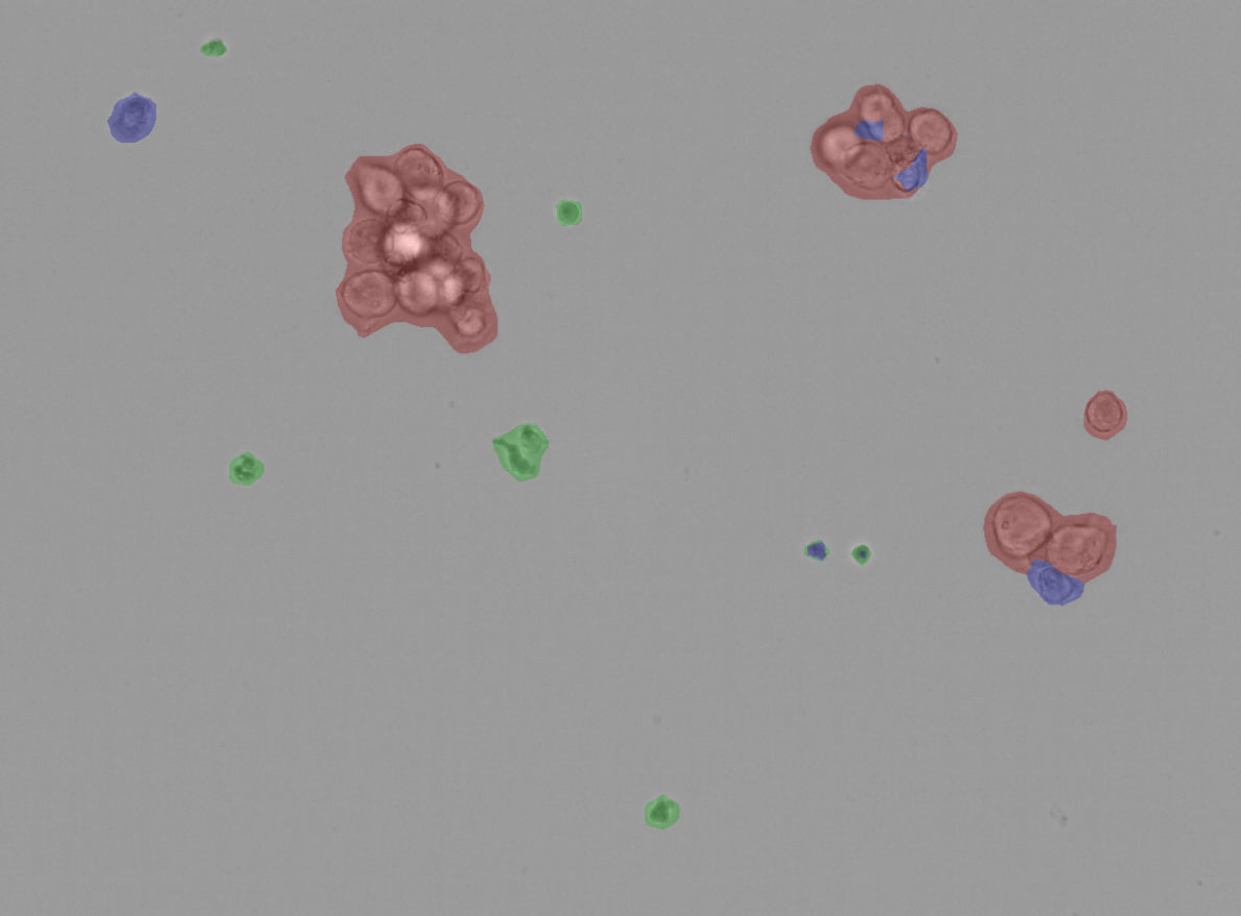} \vspace{-1.7cm} \\(4) \vspace{1.3cm}\\&\includegraphics[width=0.29\textwidth]{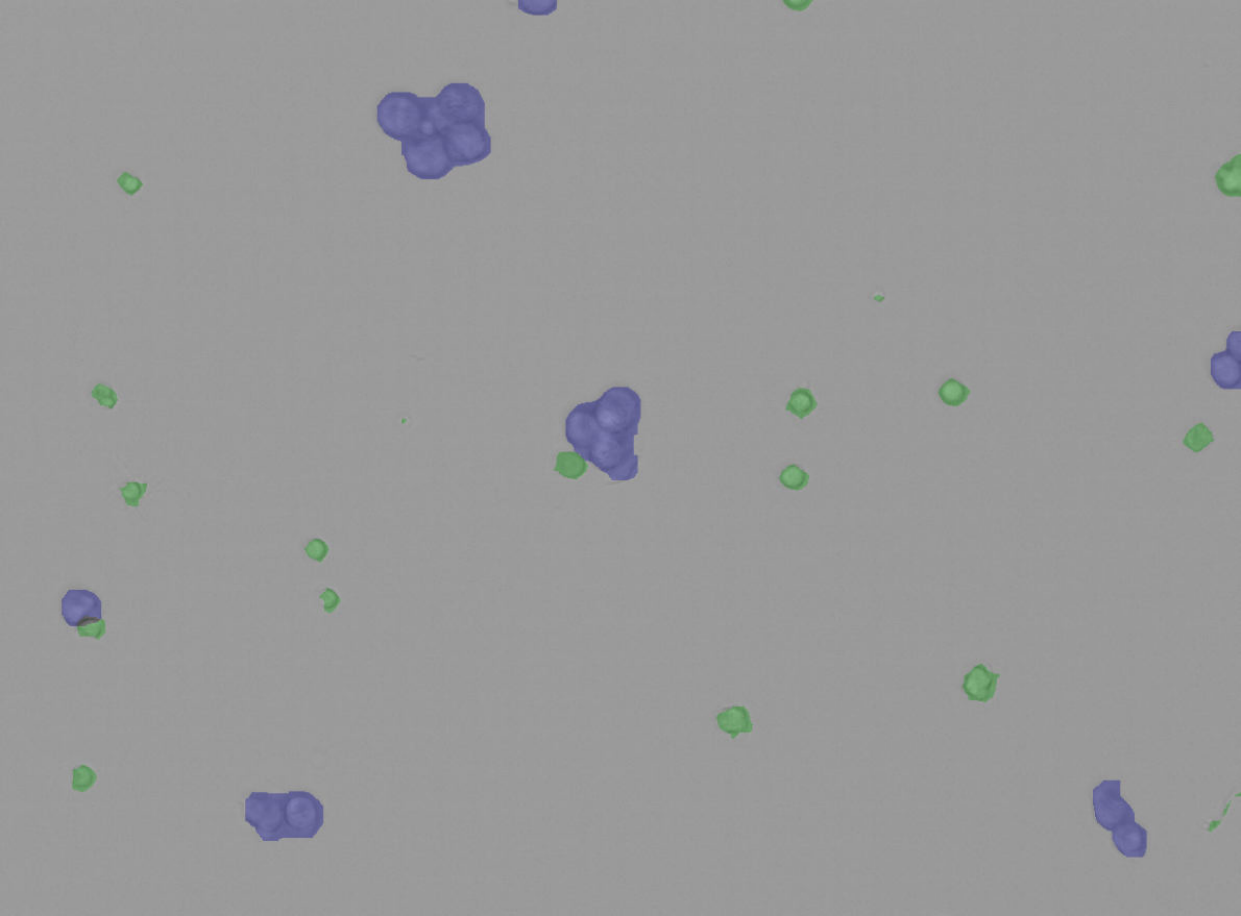}& \includegraphics[width=0.29\textwidth]{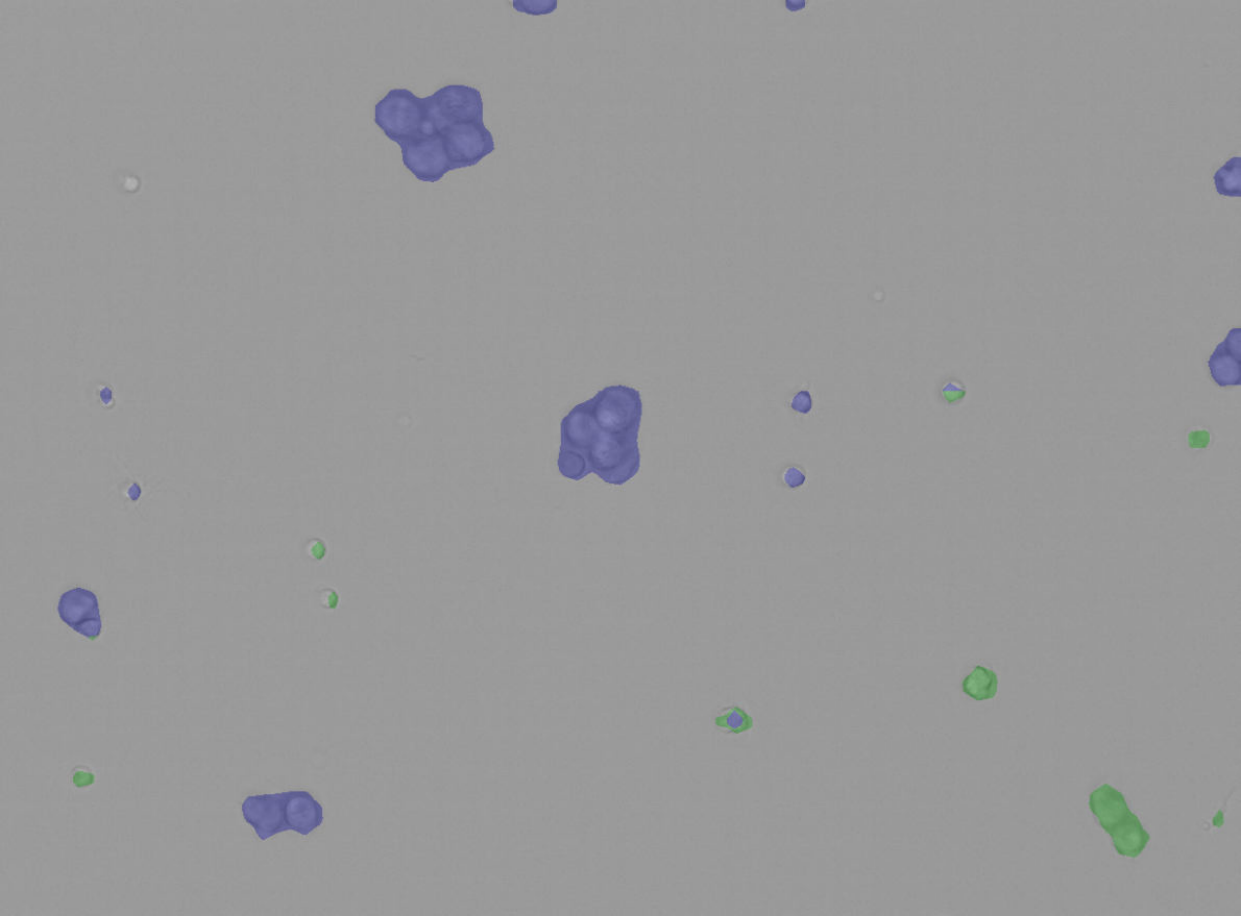}&
    \includegraphics[width=0.29\textwidth]{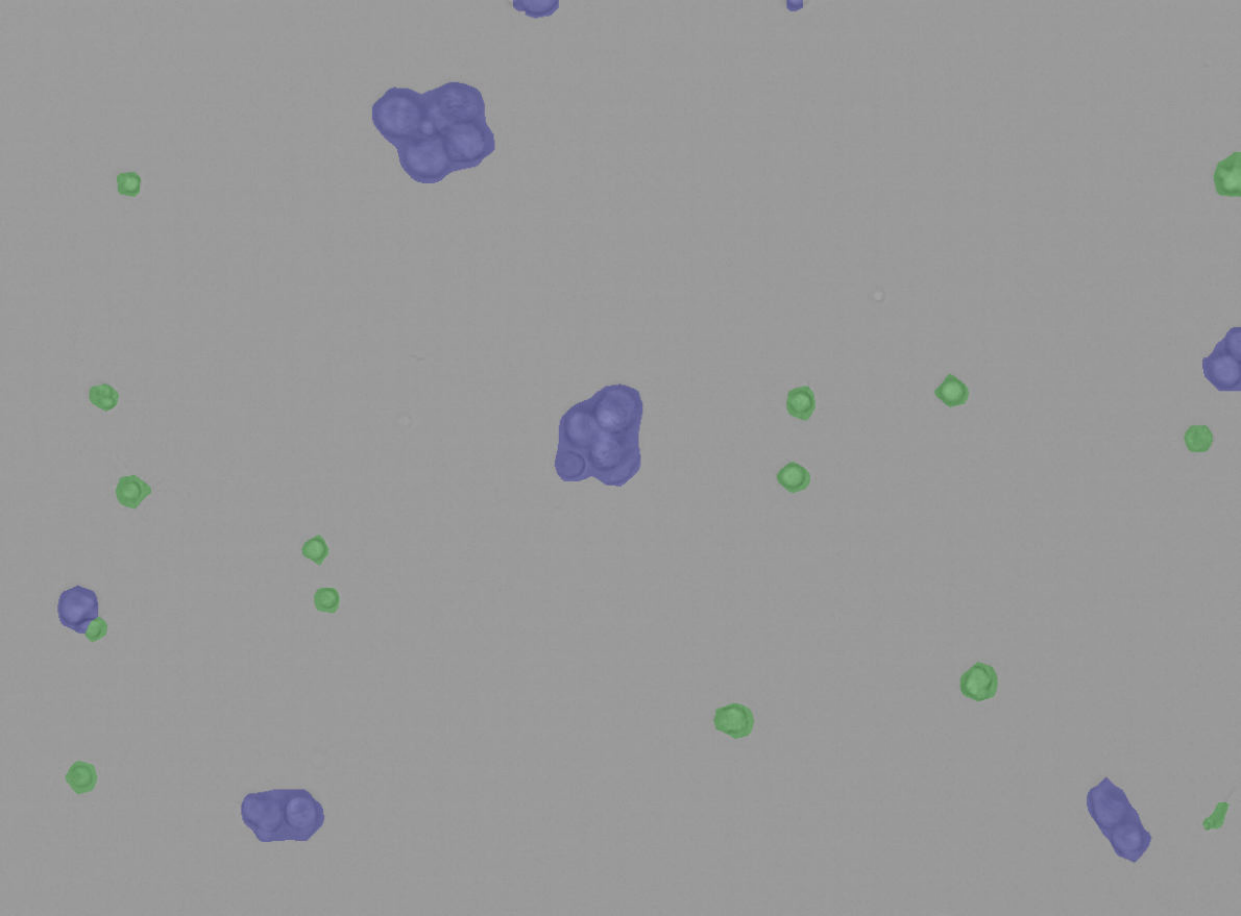} \vspace{-1.7cm} \\(5) \vspace{1.3cm}\\&\includegraphics[width=0.29\textwidth]{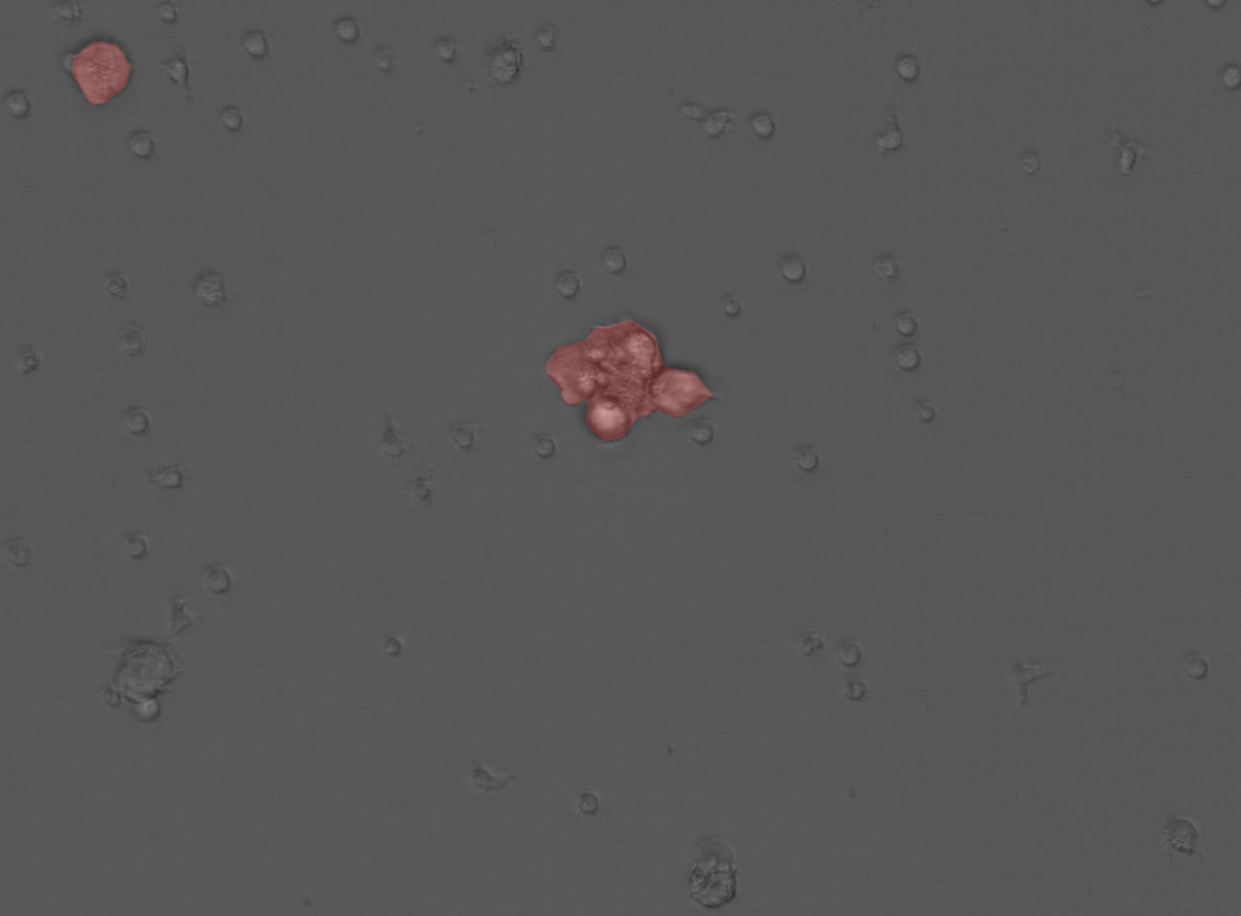}& \includegraphics[width=0.29\textwidth]{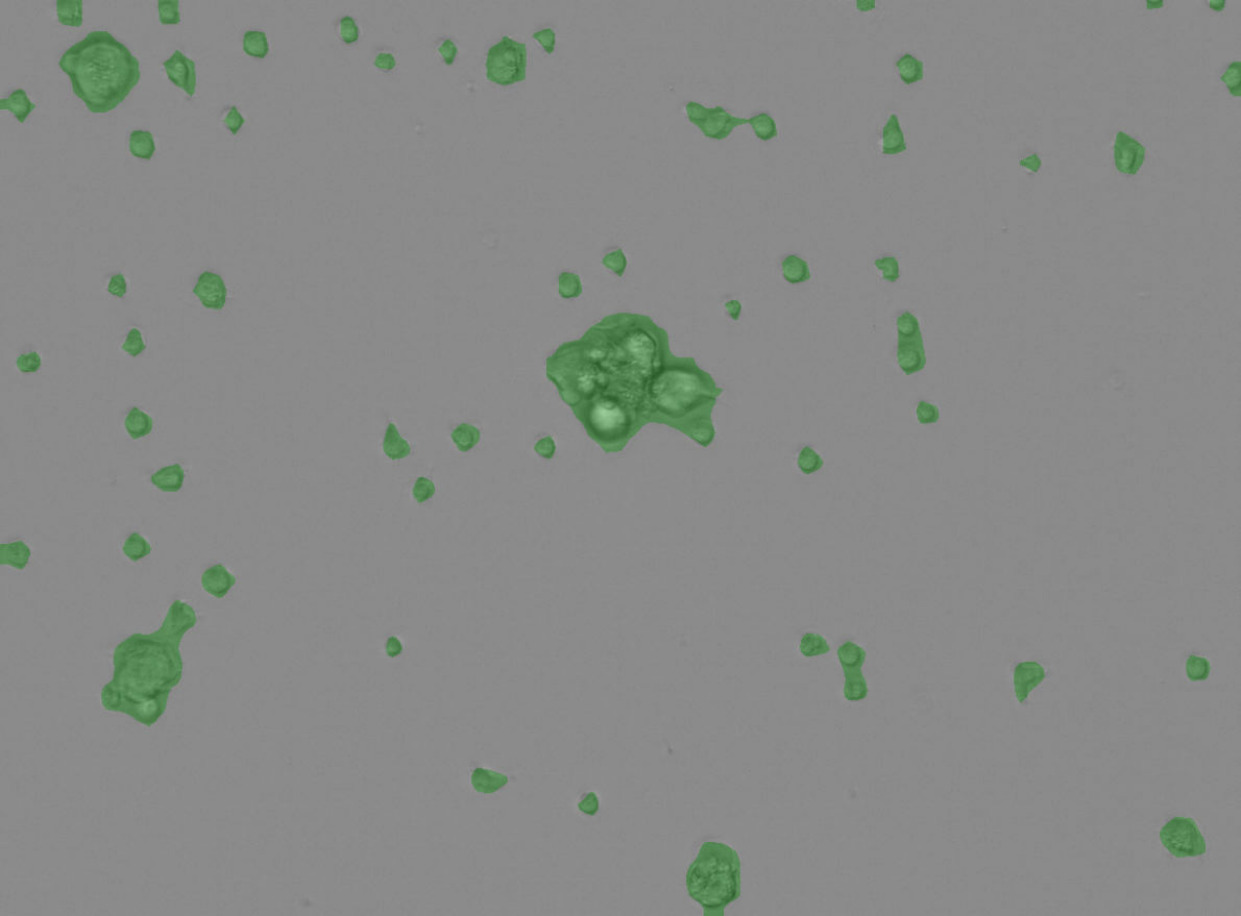}&
    \includegraphics[width=0.29\textwidth]{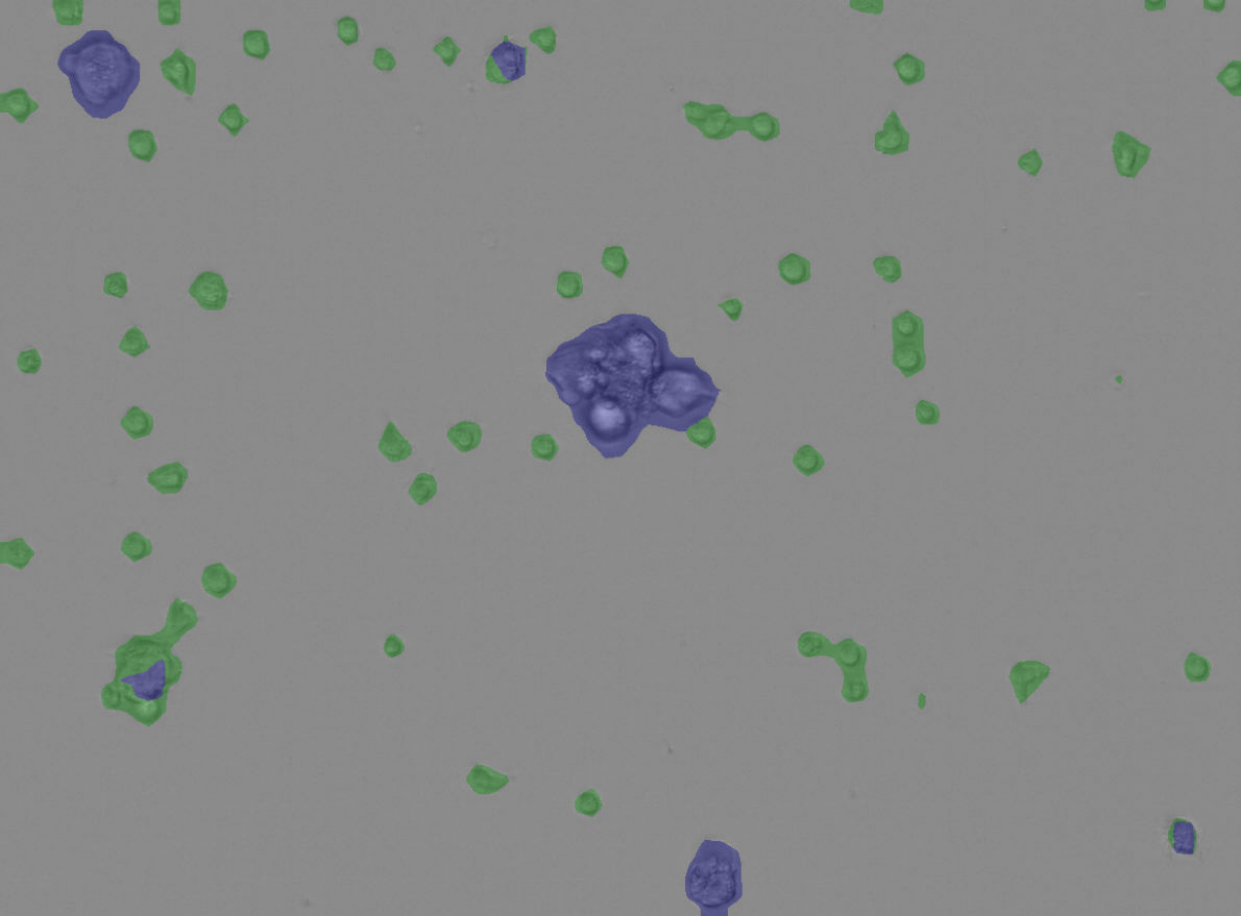} \vspace{-1.7cm} \\(6) \vspace{1.3cm}\\&\includegraphics[width=0.29\textwidth]{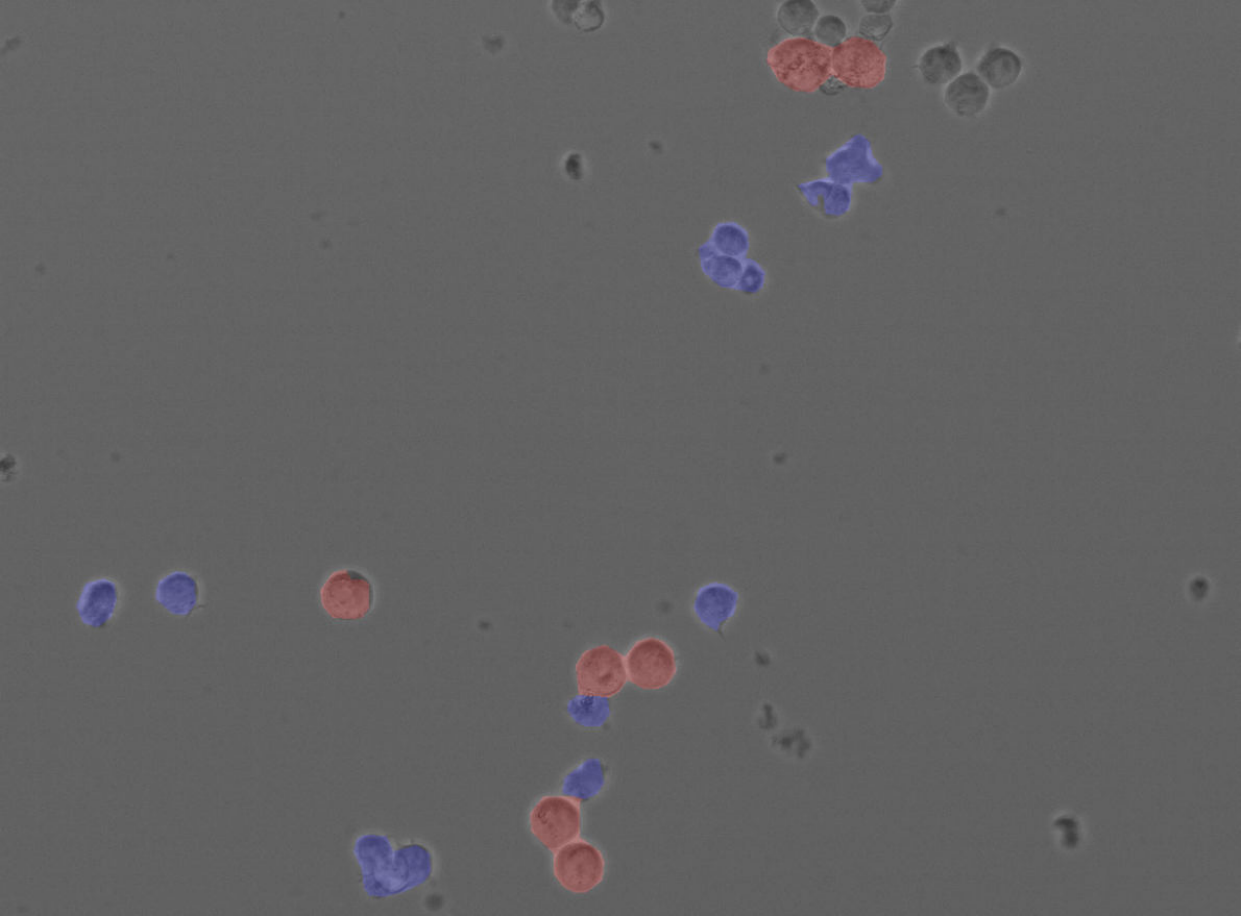}& \includegraphics[width=0.29\textwidth]{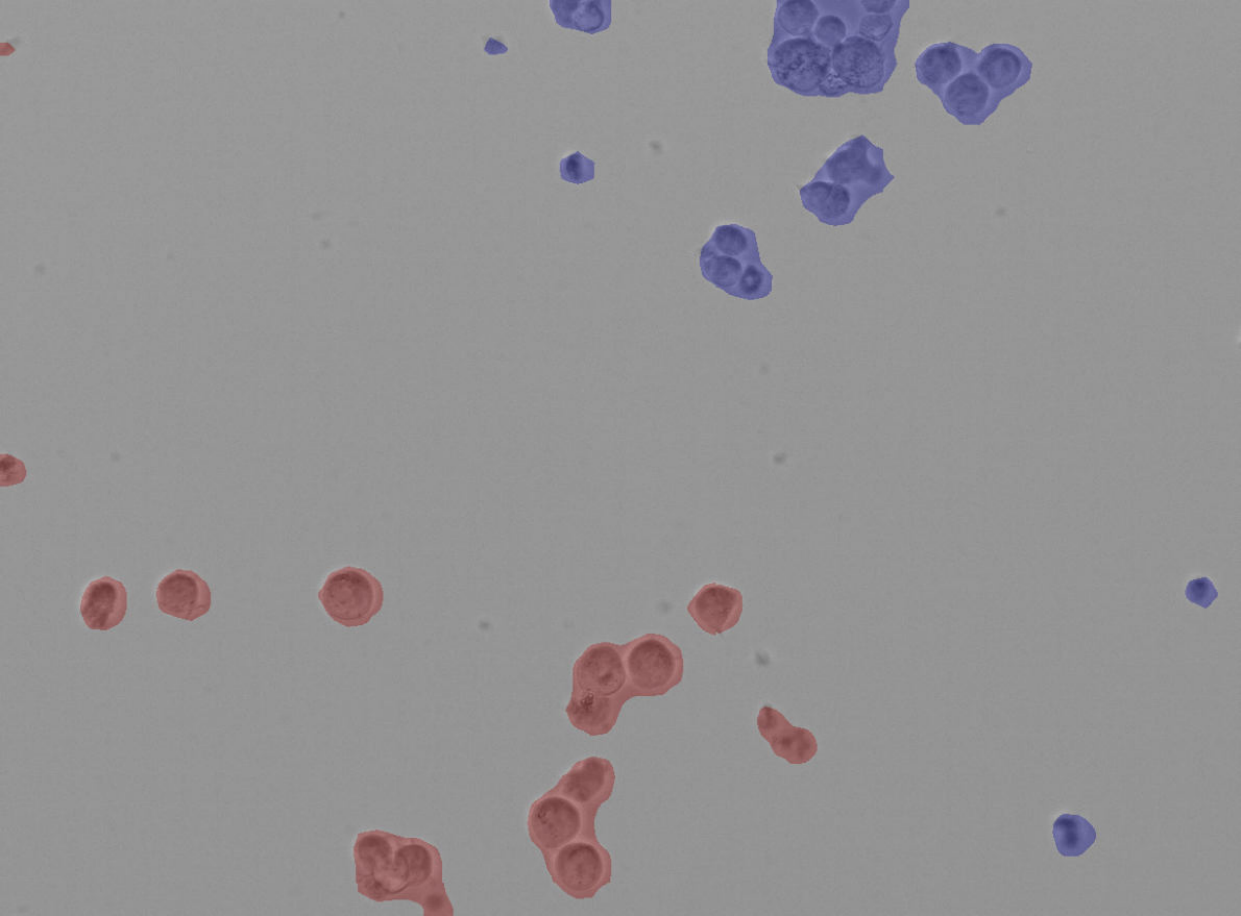}&
    \includegraphics[width=0.29\textwidth]{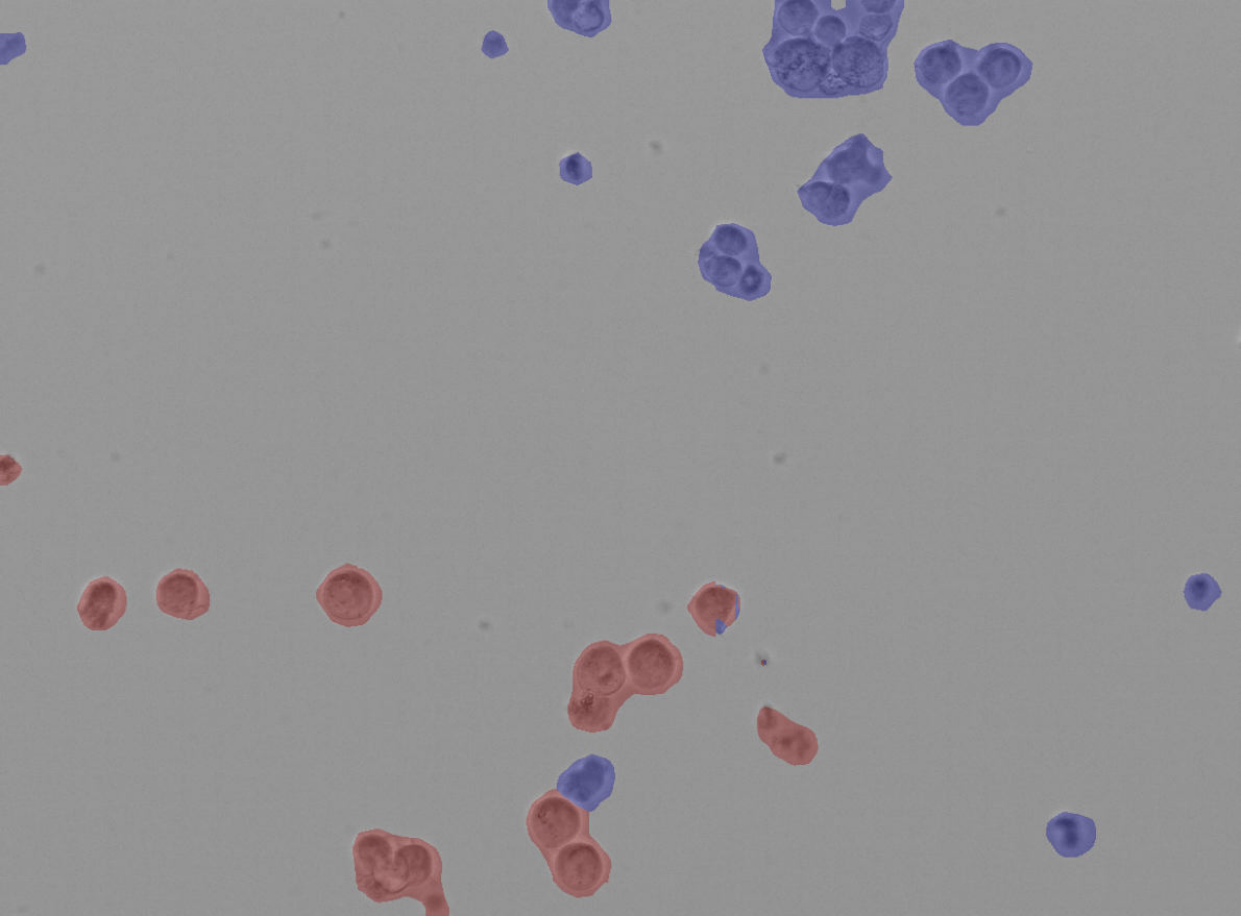} \vspace{-1.7cm} \\(7) \vspace{1.3cm}\\
	\end{tabular}
 \caption{Qualitative examples. Jurkats in blue, K562s in red and PBMCs in green.}
 \label{fig:qualitativeex}
 \end{figure}

Qualitative examples are shown in \Cref{fig:qualitativeex}. The baseline's tendency to assign the same label to all foreground pixels in its field of view can be seen in each example. Moreover, we observe that this bias is considerably reduced by Cellmixer, but not completely suppressed (see (7)). Two examples of the most challenging samples containing very dense cell clusters are presented in (4) and (6)

\end{document}